\documentclass[runningheads]{llncs}
\usepackage{graphicx}
\usepackage{amsmath,amssymb} 
\usepackage{color}
\usepackage[width=122mm,left=12mm,paperwidth=146mm,height=193mm,top=12mm,paperheight=217mm]{geometry}

\usepackage{empheq}
\usepackage{bm}
\usepackage{centernot}
\usepackage{soul}
\usepackage{parskip}
\usepackage[table]{xcolor}
\usepackage[subtle, title=normal, bibliography=normal]{savetrees}

\newcommand{\bb}[1]{\textbf{#1}}
\newcommand{\mbb}[1]{\mathbb{#1}}
\newcommand{\mc}[1]{\mathcal{#1}}

\usepackage{cancel}

\newcommand{\lb}{\left [}
\newcommand{\rb}{\right ]}

\newcommand{\etal}{et al.\,}
\newcommand{\yc}{\cellcolor{yellow!50}}

\newtheorem{mydef}{Definition}

\begin{document}
\pagestyle{headings}
\mainmatter
\def\ECCV18SubNumber{1388}  

\title{CubeNet: Equivariance to 3D Rotation \\and Translation} 

\titlerunning{CubeNet: Equivariance to 3D Rotation and Translation}

\authorrunning{D. Worrall and G. Brostow}

\author{Daniel Worrall and Gabriel Brostow}
\institute{Computer Science Department, University College London, UK\\
\email{\{d.worrall,g.brostow\}@cs.ucl.ac.uk}}

\maketitle

\begin{abstract}
3D Convolutional Neural Networks are sensitive to transformations applied to their input. This is a problem because a voxelized version of a 3D object, and its rotated clone, will look unrelated to each other after passing through to the last layer of a network. Instead, an idealized model would preserve a meaningful representation of the voxelized object, while explaining the pose-difference between the two inputs. An equivariant representation vector has two components: the invariant identity part, and a discernable encoding of the transformation. Models that can't explain pose-differences risk ``diluting'' the representation, in pursuit of optimizing a classification or regression loss function. 

We introduce a Group Convolutional Neural Network with linear equivariance to translations \emph{and} right angle rotations in three dimensions. We call this network \emph{CubeNet}, reflecting its cube-like symmetry. By construction, this network helps preserve a 3D shape's global and local signature, as it is transformed through successive layers. We apply this network to a variety of 3D inference problems, achieving state-of-the-art on the ModelNet10 classification challenge, and comparable performance on the ISBI 2012 Connectome Segmentation Benchmark. To the best of our knowledge, this is the first 3D rotation equivariant CNN for voxel representations.
\end{abstract}

\section{Introduction}
Convolutional neural networks (CNNs) are the go-to model for most prediction-based computer vision problems. However, most popularized CNNs are treated as black-boxes, lacking interpretability and simple properties concerning the data domains they act on. For instance, in 3D object recognition, we know that object categories are \emph{invariant} to object pose, but convolutional neural network filters are orientation, scale, reflection, and parity (point reflection) selective. This means that every activation in any intermediate layer is sensitive to local pose, and ultimately the global output of the network is too. A simple solution to obtain this sought-after invariance is to augment the input data with transformed copies, spanning all possible variations, to which we seek to be invariant \cite{BarnardC91}. This method is simple and effective, but relies on an efficient and realistic data augmentation pipeline. There is also the argument, why should we bother learning these invariances, if we can enforce them \emph{a priori}? If successful, we would not need as much training data \cite{CohenW16,WorrallGTB17}. Indeed, convolutional neural networks already have i) filter locality and ii) translational weight-tying built directly into their architectures, which arguably could be learned using a multilayer perceptron with a enough computational budget and training data.

We introduce a CNN architecture, which is \emph{linearly equivariant} (a generalization of invariance defined in the next section) to 3D rotations about patch centers. To the best of our knowledge, this paper provides the first example of a CNN with linear equivariance to 3D rotations and 3D translations of voxelized data. By exploiting the symmetries of the classification task, we are able to reduce the number of trainable parameters using judicious weight tying. We also need less training and test time data augmentation, since some aspects of 3D geometry are already `hard-baked' into the network. We demonstrate state-of-the-art and comparable performance on i) the ModelNet10 classification challenge, which is a standard 3D classification benchmark task, and ii) the ISBI 2012 connectome segmentation benchmark, which is a 3D anisotropic boundary segmentation problem. We will release our code as per usual.

\section{Background}
For completeness, we set out our terminology and definitions. We outline definitions of linear equivariance, invariance, groups, and convolution, and then combine these ideas into the group convolution, which is the workhorse of the paper. These definitions are not our contribution and can be found in textbooks such as \cite{Chirikjian00}, but we have tried to standardize them and simplify notation.

\begin{mydef}[Equivariance]
Consider a set of transformations $G$, where individual transformations are indexed as $g\in G$. Consider also a function or feature map $\bm{\Phi}: \mc{X} \to \mc{Y}$ mapping inputs $\bb{x}\in\mc{X}$ to outputs $\bb{y}\in \mc{Y}$. Transformations can be applied to any $\bb{x} \in \mc{X}$ using the operator $\mc{T}_g^{\mc{X}}: \mc{X} \to \mc{X}$, so that $\bb{x} \mapsto \mc{T}_g^{\mc{X}}[\bb{x}]$. The same can be done for the outputs with $\bb{y} \mapsto \mc{T}_g^{\mc{Y}}[\bb{y}]$. We say that $\bm{\Phi}$ is \emph{equivariant} to G if
\begin{align}
	\bm{\Phi}(\mc{T}_g^{\mc{X}}[\bb{x}]) = \mc{T}_g^{\mc{Y}}[\bm{\Phi}(\bb{x})], \qquad \forall g \in G. \label{eq:equivariance}
\end{align}
\end{mydef}
Since $\mc{T}_g^{\mc{X}}$ and $\mc{T}_g^{\mc{Y}}$ are related via (\ref{eq:equivariance}), they are essentially different \emph{representations} of the same transformation. Due to this connection, it is customary to drop the $\mc{T}_g^\bullet$ notation and write 
\begin{align}
	\bm{\Phi}(g\bb{x}) = g\bm{\Phi}(\bb{x}).
\end{align}

Equivariance is important, because it highlights an explicit relationship between input transformations and feature-space transformations, which in the context of deep learning is not well-understood. An example of an equivariant task is pose-detection, where $g$ represents the sought-after pose. The kind of equivariant feature maps, we are interested in, are those where $\mc{T}^{\mc{X}}$ and $\mc{T}^\mc{Y}$ are linear. Such feature maps are known as \emph{linearly equivariant}. A special case of equivariance is \emph{invariance}, where we have 
\begin{align}
	\bm{\Phi}(\bb{x}) = \bm{\Phi}(g\bb{x}),
\end{align}
that is, the feature-space transformation is just the identity. An example of an invariant task is object classification. Note when we use the term equivariant in the rest of the paper, we will generally refer to non-invariance.

\subsubsection{Groups}
Invertible transformations are members of a class of mathematical objects called \emph{groups}. Groups are a mathematical abstraction, which are used to describe the compositional structure of mathematical operators, such as transformations. Groups have four main properties: for group elements $f,g,h\in G$ 
\begin{enumerate} 
\item \bb{closure}: chained transformations are transformations, e.g.\: $fg \in G$
\item \bb{associativity}: f(gh) = (fg)h = fgh
\item \bb{identity}: there exists a transformation $e\in G$ (sometimes written $\bb{0}$) such that $eg = ge = g, \forall g\in G$
\item \bb{invertibility}: every transformation $g$ has an inverse $g^{-1}$, so $gg^{-1} = g^{-1}g = e$. Rotations and translations are both examples of groups.
\end{enumerate}

\subsubsection{Convolution}
\begin{figure}
	\centering
	\includegraphics[width=0.75\textwidth]{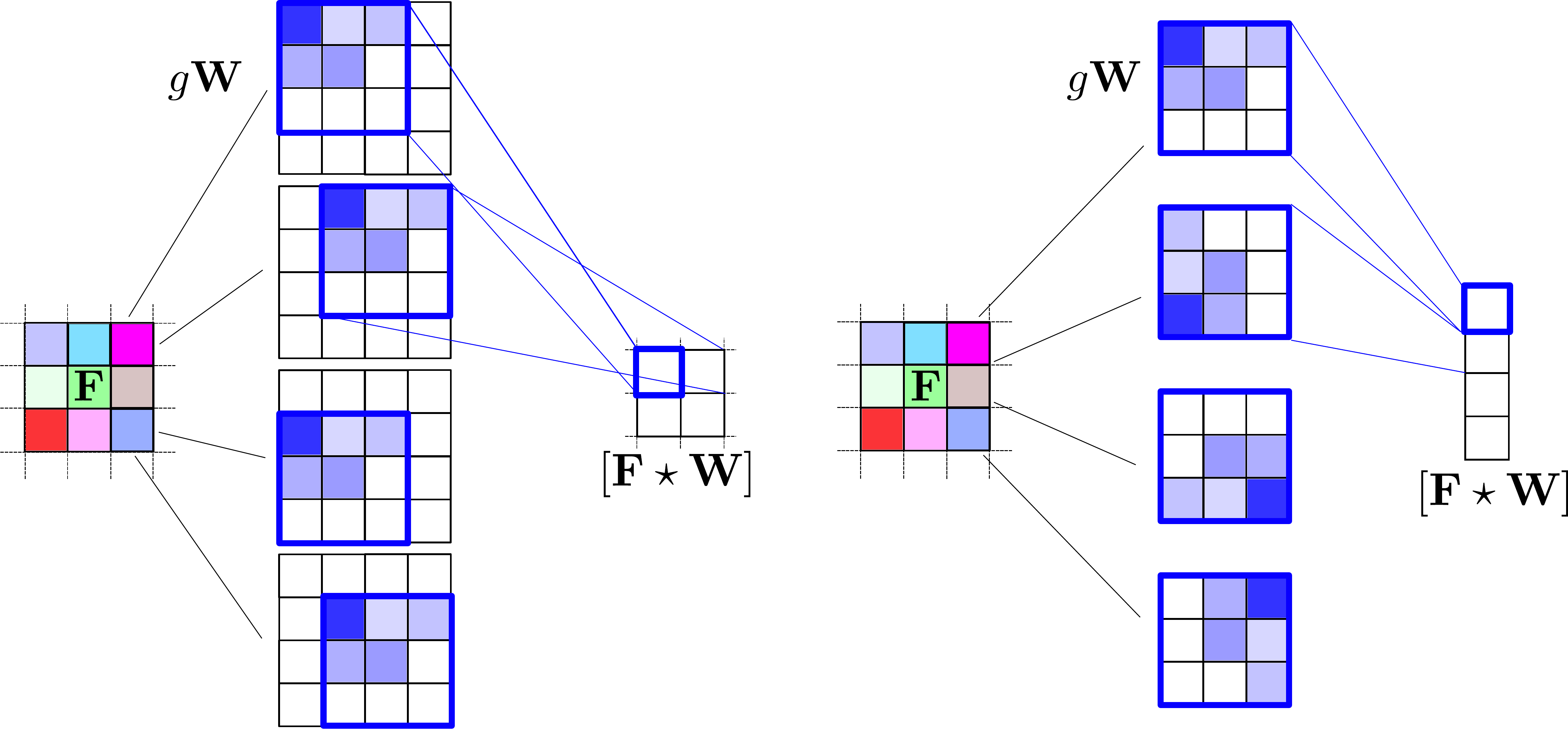}
    \caption{(Best viewed in color) On the left we show the standard 2D convolution of Equation \ref{eq:convolution} between a sliding filter $\bb{W}$ and an input patch $\bb{F}$. On the right we show the 2D right-angle rotation convolution (called $Z_4$-convolution) acting on an input where $G=\mbb{Z}^2$.}
    \label{fig:convolutions}
\end{figure}

The fundamental operation in convolutional neural networks is the convolution\footnote{Technically CNNs perform cross-correlation, but we stick with the term `convolution' to remain in sync with the literature.} $\star$. In 3D, convolution is the inner product of a \emph{filter} $\bb{W}\in\mbb{R}^{h\times w \times d}$ with patches extracted from an \emph{activation tensor} or \emph{feature map} $\bb{F}\in \mbb{R}^{H \times W \times D}$ where $h,w,d,H,W,D$ are the \bb{h}eight, \bb{w}idth, and \bb{d}epth of the filter/activations respectively. The method of patch extraction is usually a translationally sliding window. So given a filter $\bb{W}$, the translated version is $g\bb{W}$, such that
\begin{align}
	[ \bb{F} \star \bb{W} ]_g = \sum_{\bb{x}\in\mbb{Z}^3} [g\bb{W}]_{\bb{x}} \bb{F}_{\bb{x}} = \sum_{\bb{x}\in\mbb{Z}^3} \bb{W}_{g^{-1}\bb{x}} \bb{F}_{\bb{x}}; \label{eq:convolution}
\end{align}
where to index elements of the filters/activations we have used the multi-index notation $\bb{W}_\bb{x} := \bb{W}_{x,y,z}$ for $\bb{x} = [x,y,z]^\top \in\mbb{Z}^3$, and so in this example $\bb{W}_{g^{-1}\bb{x}} = \bb{W}_{x - g_x, y - g_y, z - g_z}$ for voxel-wise translation in 3D by $g = [g_x, g_y, g_z]^\top$. This sliding-window interpretation of convolution can be viewed as applying the same filter to different local regions of the inputs. Note that in reality, since the feature map is zero outside of a a certain neighborhood, we need not sum over all $\mbb{Z}^3$. Note also how the output of the convolution is indexed by the transformation parameter $g$; that is, the $g$\textsuperscript{th} activation corresponds to the response of a $g$-shifted filter $g\bb{W}$. We have used the notation $[\bb{F}\star\bb{W}]_g$ to emphasize that $[\bb{F}\star\bb{W}]$ is an indexable object like $\bb{W}$ or $\bb{F}$, and it can be viewed as a vector (see Figure \ref{fig:convolutions}). CNNs usually have multiple \emph{channels} $k$ per activation tensor, so in general we really have
\begin{align}
	[\bb{F} \star \bb{W}]^k_g = \sum_{i=1}^I \sum_{\bb{x}\in\mbb{Z}^3} [g\bb{W}]_{\bb{x}}^{ik} \bb{F}_{\bb{x}}^i, \label{eq:multi_channel_convolution}
\end{align}
where the dummy index $i$ is over input channels with output channel $k$.

One can show (c.f.\, \cite{CohenW16,CohenW16a} and Equations \ref{eq:perm1} \& \ref{eq:perm2}) that the standard translational convolution is equivariant to translations; that is, translations of the input to the convolution result in translations in the feature space representation $[\bb{F} \star \bb{W}]$. The extension of this translational equivariance to other groups of transformation is embodied in the \emph{group convolution} \cite{CohenW16}, which we show next. This has been proven \cite{KondorT18} to be the only operator which is equivariant to (compact) group-structured transformations.
\begin{mydef}[Group Convolution]
A group convolution between a filter $\bb{W}$ and a single-channel feature map $\bb{F}$ over a group of transformations $G$ is
\begin{align}
	[\bb{F} \star \bb{W}]_g = \sum_{h \in G} [g\bb{W}]_{h} \bb{F}_{h} = \sum_{h \in G} \bb{W}_{g^{-1}h} \bb{F}_{h}. \label{eq:group_convolution}
\end{align}
The extension to multichannel activations parallels Equation (\ref{eq:multi_channel_convolution}).
\end{mydef}
We see that the main difference between the standard convolution of Equation \ref{eq:convolution} and the group convolution of Equation \ref{eq:group_convolution} is that we have replaced the domain of summation from $\mbb{Z}^3$ to the group $G$. So the sliding inner product could generalize to a sliding-and-rotating inner-product, or sliding-and-flipping inner product, or even sliding-and-scaling inner product depending on the choice of group $G$. A simple example is shown in Figure \ref{fig:convolutions}, where we show a 2D translational convolution and a first layer 2D right-angle rotational convolution (called $Z_4$-convolution). In this example, the domain of the $Z_4$-convolution is $G=\mbb{Z}^2$, the standard 2D image domain, but the output is over the group of four 2D rotations, $Z_4$. This amounts to taking an inner product of the kernel $\bb{W}$ rotated four times, with each individual response being stacked into a vector. If we were to then convolve a kernel over the response of this first $Z_4$-convolution, the domain of that convolution would be $G=Z_4$.

In this paper, we are interested in the group of 3D roto-translations. The group convolution for this group will involve us convolving an activation tensor with rotated and shifted copies of a filter $[g\bb{W}]_{\bb{x}} = \bb{W}_{g^{-1}\bb{x}} = \bb{W}_{\bb{R}_g^{-1}\bb{x} - \bb{z}_g}$, where $\bb{R}_g$ is a 3D rotation matrix and $\bb{z}_g$ is a translational offset.

\section{Related Work}
Recently there has been an explosion of interest into CNNs with predefined transformation equivariances, beyond translation \cite{CohenW16,WorrallGTB17,CohenW16a,KondorT18,FaselG06,SifreM13,OyallonM15,DielemanFK16,GonzalezVT16,LaptevSBP16,GonzalezVKT17,HenriquesV17,Esteves17,ZhouYQJ17,WeilerHS2017,JacobsenBS17,JacobsenOMS17} \cite{ThomasSKYLKR18,LiYLC17,Kondor18,CohenGKW18}. However, with the exception of Cohen and Welling \cite{CohenGKW18} (projections on sphere), Kondor \cite{Kondor18} (point clouds), and Thomas \etal \cite{ThomasSKYLKR18} (point clouds), these have mainly focused on the 2D scenario. There are also examples of CNNs, which have explicit regularization to learn equivariance \cite{SimardVLD91,KulkarniWKT15,WorrallGTB17b,SabourFH17}. To the best of our knowledge, \emph{we are the first to develop a 3D rotation equivariant CNN architecture for voxelized data.}

\paragraph{Handcrafted equivariance}
There are many computer vision models that exhibit equivariance properties. Perhaps the first notable instance is the scale-space \cite{CrowleyP84}, which specifically displays equivariance to isotropic scale, later extended to affine equivariance by Lindeberg \cite{Lindeberg11}. In the presence of continuous transformations, Freeman and Adelson famously \cite{FreemanA91} (and less famously  Lenz \cite{Lenz90}), shored up the theory of \emph{steerable filters}, which are a set of bandlimited linear filters $\bb{w}_\theta \in \mbb{R}^{H\times W}$, which can be synthesized \emph{exactly} at any rotation $\theta$ as a \emph{finite} linear combination of basis filters
\begin{align}
	w_\theta(\bb{x}) = \sum_{n=1}^N \alpha_n(\theta) \phi_n(\bb{x}). \label{eq:steerable_filters}
\end{align}
These are attractive because their expressiveness is controlled by the number of coefficients $N$, rather than the spatial size of the filter. These have been applied to scale-spaces/pyramids in Simoncelli~\etal~\cite{SimoncelliFAH92}, and have been placed on firm theoretical ground by Teo~\cite{Teo98} in his PhD thesis. It has also been shown that for certain transformations, such as scalings (or more generally non-compact groups), exact steering is only possible if $N = \infty$. In this case, Perona \cite{Perona91} showed that he could approximate Equation \ref{eq:steerable_filters} using an SVD formulation. Like our method, all these works display handcrafted linear equivariance to a predefined set of transformations.

\paragraph{2D Rotation Invariant Neural Networks}
For CNNs, as mentioned, most works have focussed on 2D rotations. Fasel \& Gatica-Perez \cite{FaselG06}, Laptev \etal \cite{LaptevSBP16}, and Gonzalez \etal \cite{GonzalezVT16} average classifier predictions on multiple rotated copies of an input. Sifre \& Mallat \cite{SifreM13} and Oyallon \& Mallat \cite{OyallonM15} use a scattering network \cite{BrunaM13} for roto-translation invariant classification. Every layer of these networks is locally (patch-wise) rotation invariant, performing a pre-determined wavelet transform averaging responses over rotation. Cotter and Kingsbury \cite{CotterK17} recently suggested, however, that these networks lack discriminativeness, partially from the phase removal and partially from the fact that the wavelet transforms are not optimized per-task, which our method can handle.

\paragraph{2D Rotation Equivariant Neural Networks}
Henriques \& Vedaldi \cite{HenriquesV17} and Esteves \etal \cite{Esteves17} perform a log-polar transform of the input, which converts scalings and rotations about a single point into a translation. Applying a standard translation equivariant CNN to this representation is then equivariant to rotations and scalings about the image center. This is only equivariant to global rotations, and does not generalize to 3D. For locally equivariant methods Dieleman \etal \cite{DielemanFK16} maintain multiple rotated feature maps at every layer of a network; whereas, Cohen \& Welling \cite{CohenW16} rotate the filters. In the same paper, Cohen and Welling also extended this method to finite groups and later generalized this to arbitrary compact groups in \cite{CohenW16a}. Worrall \etal \cite{WorrallGTB17} generalized the filter rotation method to continuous rotations, using circular Fourier transforms to compute continuous rotation responses with a finite number of filters. At the same time Zhou \etal \cite{ZhouYQJ17} extended the filter rotation method to non-$90^\circ$ rotations using bilinear interpolation. Gonzalez \etal \cite{GonzalezVKT17} do similar, but also pool over rotations and use a representation similar to \cite{WorrallGTB17}. Weiler \etal \cite{WeilerHS2017} so far have the best solution to rotate filters, using steerable filters to solve the interpolation problem. Our method can be seen as an instance of Cohen \& Welling \cite{CohenW16} adapted to 3D rotation and translation.

\paragraph{Deeply Learned Equivariance}
There are many papers which also focus on learning equivariance. Tangent Prop by Simard \etal \cite{SimardVLD91} is a classic example of an invariance inducing regularizer. Hinton \etal \cite{HintonKW11} introduced the transforming autoencoder to build latent spaces with equivariant structure. More recently, Worrall \etal \cite{WorrallGTB17b} extended this method by imposing explicit transformation rules on the latent space. Papers such as InfoGAN by Chen \etal \cite{ChenCDHSSA16} and the Deep Convolutional Inverse Graphics Network of Kulkarni \etal \cite{KulkarniWKT15} seek to learn equivariant structure in unsupervised fashion unsupervised. Most recently Sabour \etal \cite{SabourFH17} and Hinton \etal \cite{HintonSF18} achieved highly impressive results on the MNIST dataset with capsule networks by learning approximations to affine equivariance. While these methods are very flexible, they require lots of training data

\paragraph{3D Methods} For classification, the most straightforward CNNs operating on 3D voxel data use 3D convolutions as of Equation \ref{eq:convolution} such as Maturana \& Scherer \cite{MaturanaS15} or 3D Convolutional Deep Belief Network as in Wu \etal \cite{WuSKYZTX15}. Brock \etal \cite{BrockLRW16} take this to the extreme, designing an ensemble of six 45-layer deep inception- and resnet-style networks trained with a lot of data-augmentation and rotation averaging. Sedaghat \etal \cite{SedaghatZB16} rely less on brute force, augmenting the prediction task with orientation estimation. For 3D rotation equivariant methods, Cohen \& Welling introduce the Spherical CNN \cite{CohenW18}, which operates on images projected onto the sphere, while Kondor \cite{Kondor18} and Thomas \etal \cite{ThomasSKYLKR18} operate on point clouds. All three methods use variants of a 3D extension of Worrall \etal \cite{WorrallGTB17}, which introduced continuous rotation equivariance into CNNs, by use of the shifting property of Fourier transforms.

\section{Method}
We have introduced the concept of groups as a way to model transformations, and as a way to extend standard convolution to these transformations. Here, we demonstrate a choice of three 3D rotation groups. We select Klein's four-group to build a 3D roto-translation equivariant CNN. We then show how to apply this group in a group equivariant CNN. The overall aim is to design a network that filters a voxelized 3D input shape at discrete right-angle orientations, while producing a signature that preserves the shape's identity and respects its pose.

\subsubsection{Cube Group}
\begin{figure}
\centering

\includegraphics[width=\columnwidth]{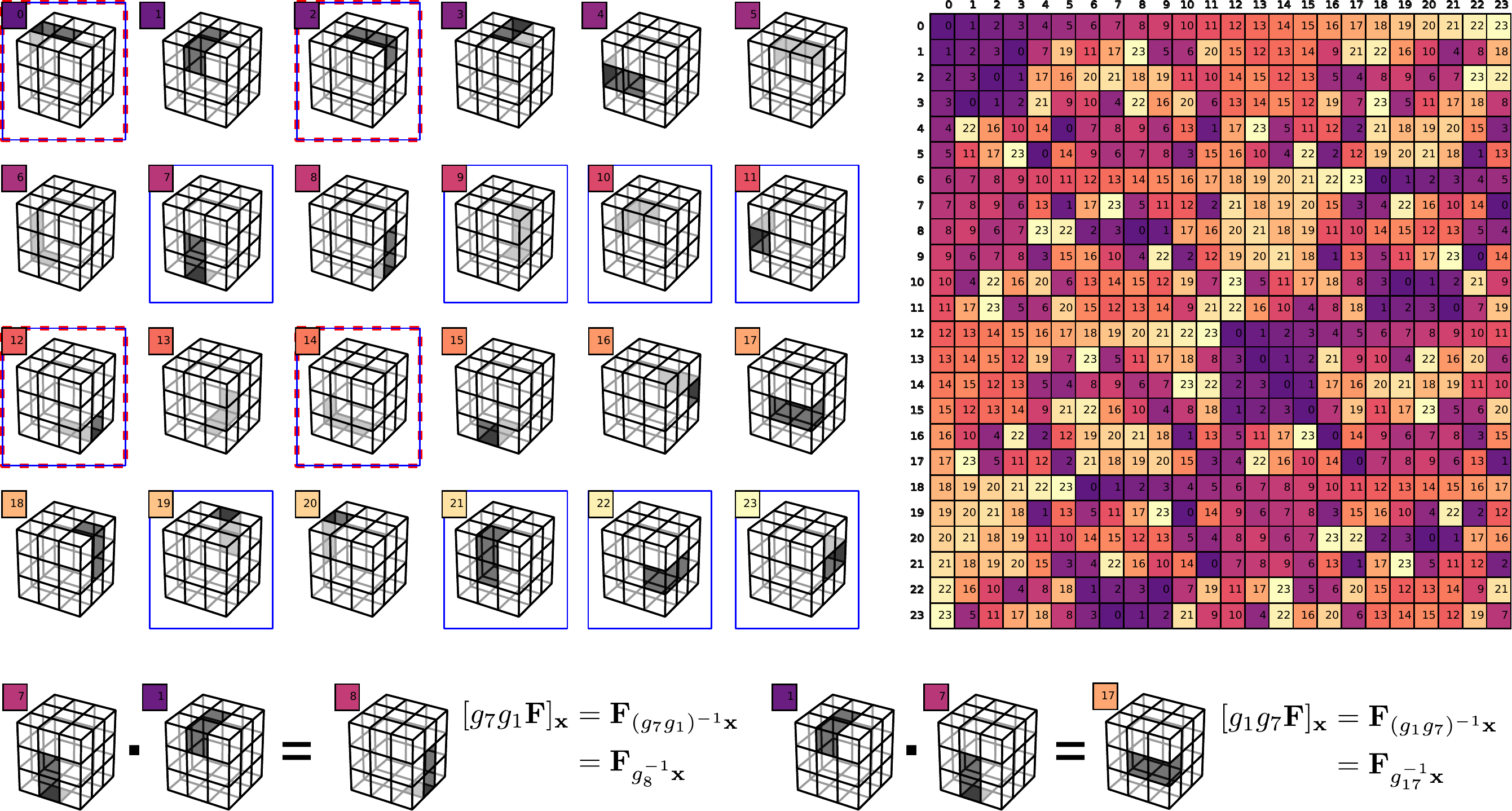}
\caption{(Best viewed in color) \textsc{Left}: The 24 rotations of the cube group $S_4$, applied to the a cube $\bb{F}_\bb{x}$ are shown. For instance, rotation $g_{22}$ applied to the cube returns $\bb{F}_{g_{22}^{-1}\bb{x}}$, shown by the \#22 in the bottom row. The 12 cubes wrapped in thin blue boxes are the rotational tetrahedral group $T_4$. The 4 cubes wrapped in thick dashed red lines are the Klein four-group $V$. \textsc{Right}: The Cayley table of the cube group, representing how rotations are composed. For instance, on the \textsc{bottom left}, we have the example of composing rotation $g_7$ with rotation $g_1$. The composition is performed by i) first applying $g_7$ to the cube to yield $\bb{F}_{g_7^{-1} \bb{x}}$ then ii) applying $g_1$ to $\bb{F}_{g_7^{-1} \bb{x}}$, returning $\bb{F}_{g_1^{-1}g_7^{-1} \bb{x}}$. The first transformation is easy to visualize - it is by $\#7$ in the grid of cubes. The transformation $g_1$ is a rotation by $90^\circ$ counter-clockwise about the vertical axis, thus for the composition we rotate $\bb{F}_{g_7^{-1}\bb{x}}$ $90^\circ$ counter-clockwise about the $z$-axis. This results in $F_{g_8^{-1}\bb{x}}$. This result is stored in the Cayley table by placing the first rotation down the left column and the second rotation along the top row. The intersection of row $\bb{7}$ with column $\bb{1}$ is the rotation $\bb{8}$. On the \textsc{bottom right}, we show the composition $g_7 g_1 = g_{17} \neq g_{8} = g_1 g_7$, demonstrating the non-commutativity property of the cube group and 3D rotations in general.}
\label{fig:cube_rotations}
\end{figure}

The set of all right-angle rotations of a cubic filter $\bb{F}_\bb{x} \in \mbb{R}^{N\times N\times N}$ forms a group. There are 24 such rotations, going by the name of the \emph{cube group}\footnote{Other names are the subgroup $O$ of the octohedral group; symmetric group $S_4$; and full tetrahedral group $T_d$.} $S_4$. Each of the 24 rotations applied to a cube is shown in Figure \ref{fig:cube_rotations}. The group is non-commutative, so $\bb{F}_{(g_1g_7)^{-1}\bb{x}} \neq \bb{F}_{(g_7g_1)^{-1}\bb{x}}$ for rotations $g_1$ and $g_7$, for example.

\subsubsection{Tetrahedral Group}
Using 24 copies of the same filter increases the computational overhead 24 times. A cheaper subsampling is the rotations of the tetrahedron. This has 12 states, and goes by the name of the \emph{rotational tetrahedral group} $T_4$. $T_4$ is formally a subgroup\footnote{A subgroup $H$ is any subset of $G$, which satisfies the four group axioms, which we introduced in the background section} of the cube group, comprised of all even rotations (i.e.\ all rotations which can be made by two $90^\circ$-rotations). It is shown as the 12 cube rotations wrapped in thin blue in Figure \ref{fig:cube_rotations}.

\subsubsection{Klein's Four-group}
The smallest subsampling of rotations, which can be seen as rotations about 3 independent axes is Klein's \emph{Vierergruppe} $V$ or \emph{four-group}. It has four rotations as can be seen in Figure \ref{fig:permutation}. This group is a subgroup of the rotational tetrahedral group and the cube group. Interestingly, it is commutative and also the smallest non-cyclic group. It is shown as the 4 rotations wrapped in dashed red in Figure \ref{fig:cube_rotations}.

\subsection{Cayley tables}
Knowing how a rotation of the input will permute the convolutional response can be figured out from the group \emph{Cayley table}. This is a multiplication table enumerating every composition of transformations. For Klein's four-group, we label the rotations as $g_0$ (the identity), $g_1$, $g_2$, \& $g_3$. The Cayley table with instructions of how to read it are given in Table \ref{tab:cayley}. The Cayley table is useful for determining how to perform the group convolution in deeper layers. We can see why this is the case because looking to the expression for the group convolution $\sum_{h\in G} \bb{W}_{g^{-1}h}\bb{F}_h$, we see a product $g^{-1}h$ in the indices of $\bb{W}$. We can use the Cayley table to ascertain the single transformation that is the result of the product. Looking closely at a Cayley table we see that all the rows/columns are permutations of one another, this will be important for understanding how input rotations affect the group-convolutional response. 

\begin{table}[t]
	\caption{The Cayley table for Klein's four-group. The product $g_2g_3$ (a $g_2$-rotation followed by a $g_3$-rotation) can be found by looking down the left column for the first transformation $g_2$, then finding the second transformation $g_3$ in the top row. The cell at the intersection of row-$g_2$ and column-$g_3$ (shaded in yellow) is $g_1$, so $g_2g_3=g_1$.}
    \label{tab:cayley}
    \centering
 	\begin{tabular}{r | c c c c}
		$\bullet$	&	$g_0$	&	$g_1$		&	$g_2$		&	\yc$g_3$ \\
      	\hline
      	$g_0$		&	$g_0$		&	$g_1$		&	$g_2$		&	\yc$g_3$ \\
      	$g_1$		&	$g_1$		&	$g_0$		&	$g_3$		&	\yc$g_2$ \\
      	\yc$g_2$	&	\yc$g_2$	&	\yc$g_3$	&	\yc$g_0$	&	\yc$g_1$ \\
      	$g_3$		&	$g_3$		&	$g_2$		&	$g_1$		&	\yc$g_0$ \\
  	\end{tabular}
\end{table}

\subsection{Discrete Group Equivariance and Permutations}
\begin{figure}
\centering
\includegraphics[width=\columnwidth]{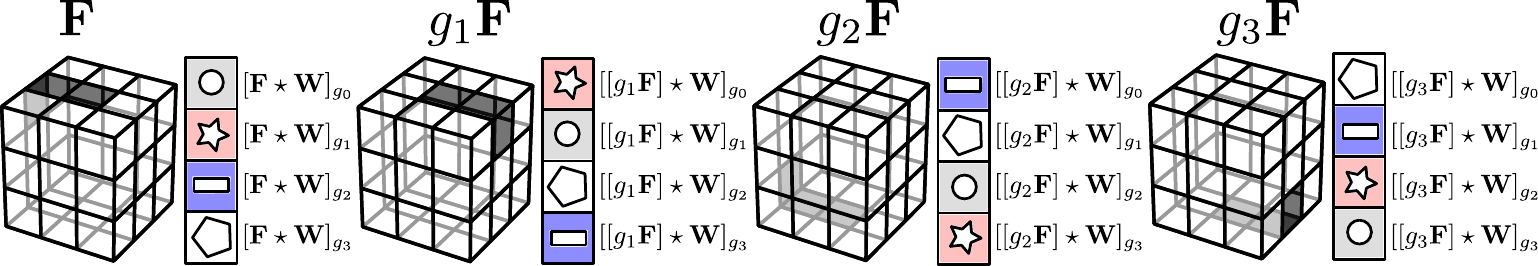}
\caption{Example of how the group convolution output permutes as a function of the input rotation. This example is for Klein's four-group $V$. Each cube represents a rotation from $V$ and a corresponding example feature vector is given with each cube.}
\label{fig:permutation}
\end{figure}

Rotating an input to a group convolution will lead to a transformation of its output. Specifically a rotation will lead to a permutation of the output, where we view the output as a vector of responses, with each dimension corresponding to a different group element/transformation $g\in G$. An example of this vectorized output can be seen in Figure \ref{fig:convolutions}. For translations the permutation is a voxel-wise shift, but for the aforementioned 3D rotations the permutations are much more complicated. If we apply a transformation $p$ to the input features $\bb{F}$, then
\begin{align}
	[[p\bb{F}] \star \bb{W}]_g &= \sum_{h \in G} [g\bb{W}]_h [p\bb{F}]_h = \sum_{h \in G} \bb{W}_{g^{-1}h} \bb{F}_{p^{-1}h} \label{eq:perm1}\\
	&= \sum_{h' \in G} \bb{W}_{g^{-1}ph'} \bb{F}_{h'} = [\bb{F} \star \bb{W}]_{p^{-1}g} = [p[\bb{F} \star \bb{W}]]_{g}. \label{eq:perm2}
\end{align}
Here we have made the substitution $h' = p^{-1}h$ and noted that $p^{-1}G = G$ for $p\in G$, where $p^{-1}G := \{p^{-1}g \mid g \in G\}$. What lines \ref{eq:perm1} and \ref{eq:perm2} say is that the output of the group convolution is permuted whenever the input $\bb{F}$ is transformed by an element of the group $G$. The specific permutation of the output depends on the specific transformation and transformation group. Thinking of $\bb{F} \star \bb{W}$ and $[p\bb{F}] \star \bb{W}$ as vectors separated by a permutation, we can write
\begin{align}
	[p\bb{F}] \star \bb{W} = p[\bb{F} \star \bb{W}] = \bb{P}_p[\bb{F} \star \bb{W}],
\end{align}
where the first equality is from Equations \ref{eq:perm1} and \ref{eq:perm2} and in the second equality we have rewritten the permutation as multiplication with the \emph{permutation matrix} $\bb{P}_p$. In fact $\bb{P}_p$ is the permutation matrix corresponding to the $p$\textsuperscript{th} column of the Cayley table. Thus we see that group convolutions are \emph{linearly} equivariant to transformations $p\in G$, as defined in Equation \ref{eq:equivariance}. We see an example of this for Klein's four-group in Figure \ref{fig:permutation}, where we have labeled the four rotations as $g_0$ (the identity), $g_1$, $g_2$, \& $g_3$.

\subsection{Implementation: Roto-translational group-convolution}
Now we show how to implement a group-convolution for a 3D roto-translation. In this example, we focus on the four-group to model rotations. A roto-translation can be synthesized from a rotation, followed by a translation. Roto-translations form a group, which can be seen as the product\footnote{Formally, this is a semi-direct product.} of $V$ and $\mbb{Z}^3$. For our purposes, it is safe to assume that we can write the elements of this producted group as $tr$ for $t\in\mbb{Z}^3$ and $r\in V$. So, 
\begin{align}
	[\bb{F} \star \bb{W}]_{tr} &= \sum_{\tau \in \mbb{Z}^3}\sum_{\rho \in V} \lb tr\bb{W} \rb_{\tau\rho} \bb{F}_{\tau\rho} = \sum_{\tau \in \mbb{Z}^3}\sum_{\rho \in V} \lb t \lb r \bb{W}\rb_{\rho} \rb_{\tau} \bb{F}_{\tau\rho}.
\end{align}
The interpretation behind this equation is as follows. First we start with a filter $\bb{W}$. $\bb{W}$ has a different value for each voxel in its receptive field, indexed by the translation variable $\tau$, and also for every input rotation $\rho$---it may be easier just to think of four 3D filters, $\bb{W}_{\rho_0}, \bb{W}_{\rho_1}, \bb{W}_{\rho_2}, \bb{W}_{\rho_3}$, one for each rotation in $V$. To convolve, we first rotate the kernel as $r\bb{W}_{\rho_\bullet}$, then we perform a translational shift $t[r\bb{W}_{\rho_\bullet}]$---this second part ends up as the standard convolution of Equation \ref{eq:convolution}, which is efficient on GPUs. The initial rotation of the filter $r\bb{W}_{\rho_\bullet}$ can be found from composing $r$ and $\rho_\bullet$ using our Cayley tables. It is the rotation needed to rotate $r$ into $\rho_\bullet$. When the input is a raw image, the input domain is just $\mbb{Z}^3$, so the rotation of $\bb{W}$ is just $r$.

\section{Experiments and results}
Here we describe two simple experiments we performed to demonstrate the effectiveness of group-convolutions on 3D voxelized data. We tested on the ModelNet10 classification challenge, which is a small 3D voxel dataset, and on the ISBI 2012 connectome segmentation challenge. In both examples, we found Klein's four-group to be the most effective group for the rotation-equivariant group-convolutions.

\subsection{ModelNet10}
The ModelNet 10 dataset \cite{WuSKYZTX15} contains 4905 CAD models from 10 categories with a train:test split of 3991:914. Each model is aligned to a canonical frame and then rotated at 12 evenly-sampled orientations about the $z$-axis. These rotated models are then voxelized to a 32x32x32 grid. We use the voxelized version of Maturana and Scherer \cite{MaturanaS15}. While the dataset consists of vertically aligned models, rotated only about the $z$-axis, we posit that local features occur at all 3D rotations, and so a Cubenet is well positioned to operate on such as dataset. We use the four-group of rotations and the rotational tetrahedral group $T_4$, since we found the cube-group too large and slow to be trained practically multiple times during a model search.

\begin{figure}
	\includegraphics[width=\columnwidth]{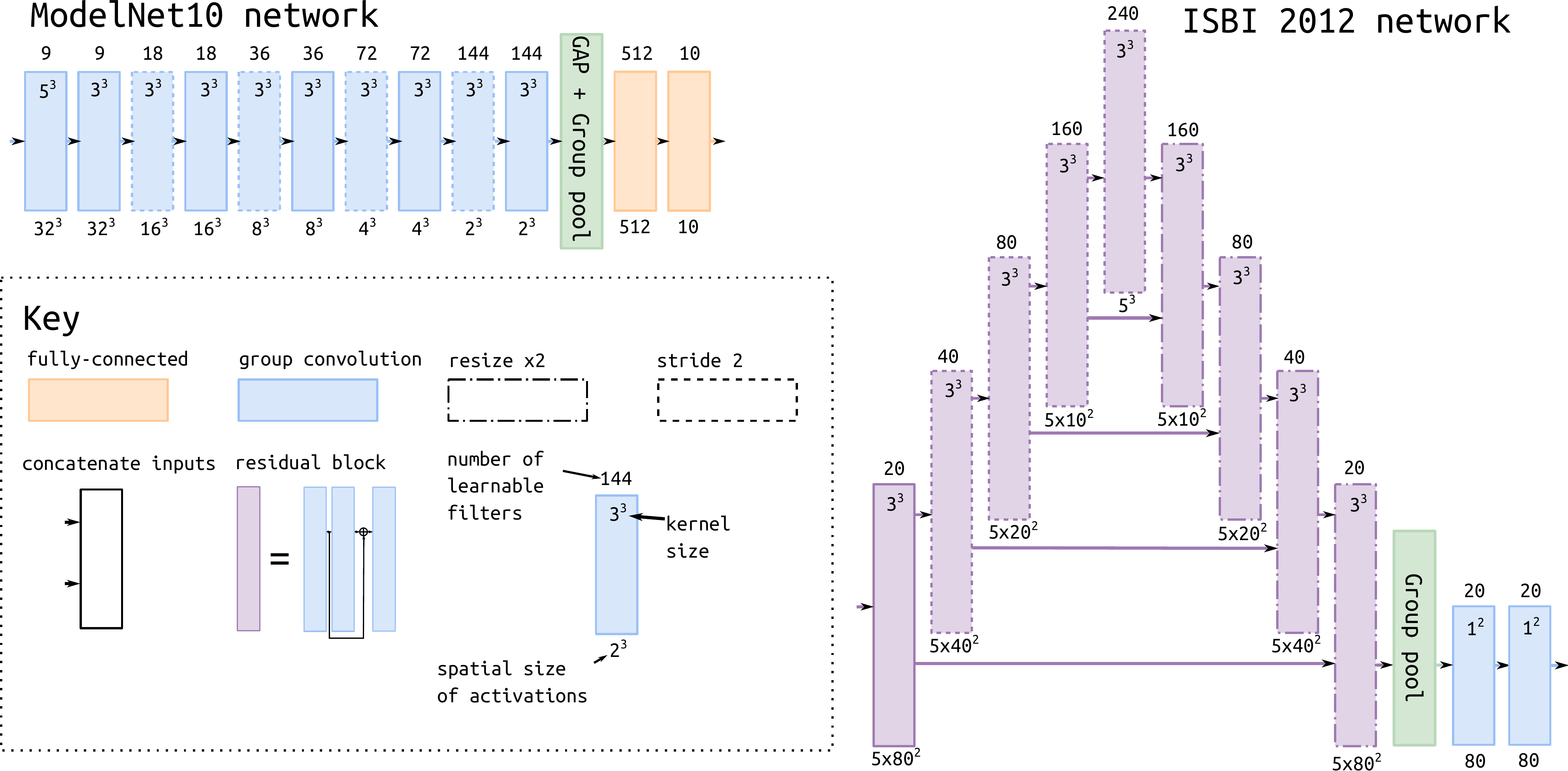}
    \caption{(Best viewed in color) The architectures used in our experiments. We use a simple VGG-like architecture for the ModelNet10 classification challenge, and a UNet/FusionNet-like architecture for the ISBI2012 boundary segmentation benchmark.}
    \label{fig:architectures}
\end{figure}

We use a simple VGG-like \cite{SimonyanZ14a} network architecture shown in Figure \ref{fig:architectures}. It consists of 10 group-convolutional layers followed by a 2-layer fully-connected network. Before every convolution, we combine multiplicative dropout with $0.1$ standard deviation on the filter tensors, and after every convolution we add batch normalization. We use ReLU nonlinearities and global average pooling before two fully-connected layers at the end of the network. The loss function is the multi-class cross-entropy. We initialize all weights using the He method \cite{HeZRS15} and train the network with ADAM stochastic gradient descent \cite{KingmaB14}, with a learning rate of 1e-3, which steps down by $1/5$ every 5 epochs for 25 epochs. 

The data augmentation is performed similar to the implementation found in Brock \cite{BrockLRW16} with 12 stratified rotations about the $z$-axis, reflections in the $x$- and $y$-axis with uniform probability and uniformly random translations of up to $\pm 4$ voxels along all three axes. We also rescale the voxel values to $\{-1,5\}$ instead of $\{0,1\}$ as in \cite{BrockLRW16}, who showed it helps with sparse voxel volumnes. We show our results in Table \ref{tab:modelnet10}. We compare the rotational tetrahedral group and the four-group models. For the four-group model, we compare the average single-view accuracy across 5 models for robustness, with rotation averaged accuracy and single-view accuracy for the best model. The single view accuracy is computed as the accuracy averaged over each of the 12 rotated test views; whereas, the rotation averaged accuracy is computed as the accuracy of the average of all 12 predictions.
\begin{table}
	\center
	\caption{Results for the ModelNet 10 benchmark. We compare against other methods which operate on a voxel-representation of the data. The only model to beat us is Brock \etal's ensemble of 6 models. If we just restrict to a single model, then we hold state-of-the-art accuracy.}
    \label{tab:modelnet10}
	\begin{tabular}{l c c}
    Method							& ModelNet10 	& \# params ($\times 10^{6}$, 2 s.f.) \\
    \hline
    3D ShapeNets \cite{WuSKYZTX15}	& 0.8354 		& 12 \\
    Xu \& Todovoric	\cite{XuT16}	& 0.8800 		& 0.080 \\
    3D-GAN \cite{WuZXFT16}			& 0.9100 		& 11 \\
    VRN	\cite{BrockLRW16}			& 0.9133 		& 18 \\
    VoxNet \cite{MaturanaS15}		& 0.9200 		& 0.92 \\
    Fusion-Net \cite{HegdeZ16}		& 0.9311 		& 120 \\
    ORION \cite{SedaghatZB16}		& 0.9380		& 0.91 \\
    \hline
    Ours $T_4$						& 0.9127		& 4.5 \\
    Our $V$ (average)				& 0.9372		& 4.5 \\
    Ours $V$ (best model single-view)    	& 0.9420	& 4.5 \\ 
    Ours $V$ (best model rotation averaged)	& \bb{0.9460}		& 4.5 \\
    \hline
    \hline
    VRN Ensemble \cite{BrockLRW16}	& 0.9714 		& 108 \\		
	\end{tabular}
\end{table}

For the single-model category, our four-group, rotation-averaged network attains state-of-the-art performance. Interestingly, our single-view result we obtain is very similar to ORION \cite{SedaghatZB16}, which introduces an orientation estimation task along with the classification. We posit that the $T_4$-model does not perform as well as the $V$-model, because increasing the number of filter copies reduces the diversity of filters, when the number of total filters (number of learnable filters times number of copies) is constrained. It is also interesting to see that rotation averaging improves performance slightly, compared to our single-view model. We suggest this is because we are averaging over rotations not covers by the four-group. Looking across the model sizes, we see that the group-convolutional models sit somewhere in the middle in terms of number of parameters. Speed-wise, we found that during development the four-group network only trained about $2\times$ slower than non-group CNNs.

\subsection{ISBI 2012 Challenge: Connectome Segmentation}
The ISBI 2012 Challenge is a volumetric boundary segmentation benchmark. The task is to segment Drosophilia ventral nerve cords from a serial-section transmission electron microscopy (EM) image \cite{Arganda-Carreras15}. The training set is a single $2\times 2 \times 1.5$ $\mu$m\textsuperscript{3} volume of anisotropic imaging resolution (high $x$-$y$ resolution, low $z$ resolution). Each voxel is $4 \times 4 \times 50$ nm\textsuperscript{3} so the full training image is $512 \times 512 \times 30$ voxels in shape. The test image is $512\times512\times30$ voxel, with withheld labels. Scoring is performed using the metrics $V_\text{rand}$ and $V_\text{info}$ described in \cite{Arganda-Carreras15}. Larger is better.

Here we are faced with two major issues, a) small dataset, b) high imaging anisotropy. We counter a) with heavy data augmentation as per \cite{QuanHJ16} and by noting that group convolutions reduce the number of trainable parameters through significant weight-tying. To counter the imaging anisotropy, we use Klein's four-group, which is not affected by stretching of one of the axes.

\begin{figure}
	\includegraphics[width=0.32\textwidth]{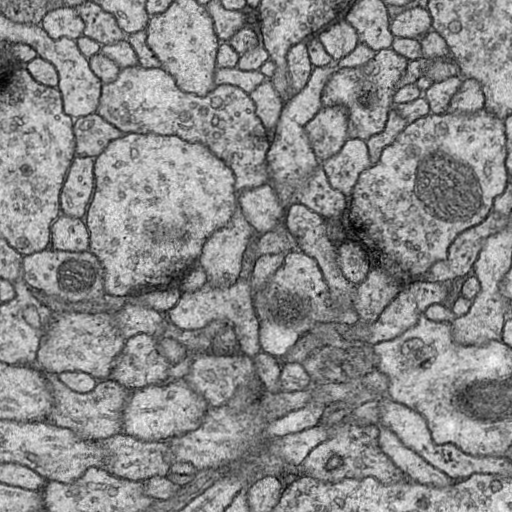}
	\includegraphics[width=0.32\textwidth]{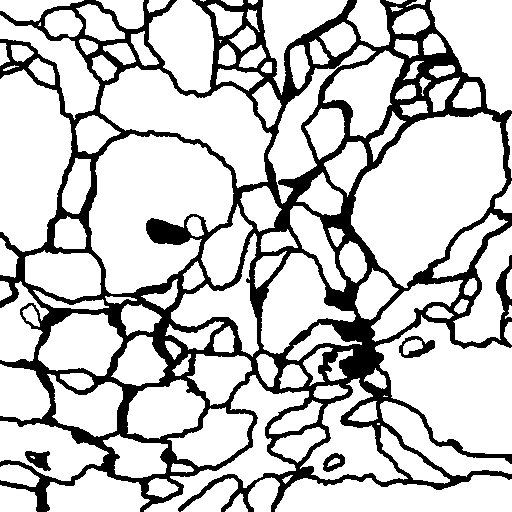}
    \includegraphics[width=0.32\textwidth]{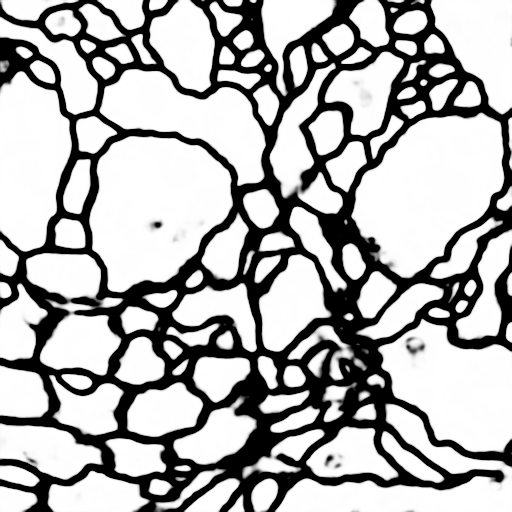}
	\caption{Examples of 2D slices from the training volume, the associated label mask, and the prediction made by our network. The original volume contains small amounts of noise and certain structures within the volume are ambiguous in nature.}
\end{figure}

Competing methods segment on a single 2D high-resolution slice at a time, but as a proof of concept we try segmentation as a 3D problem, feeding 3D image chunks into a 3D network. We use an architecture as shown in Figure \ref{fig:architectures}, based on Weiler \etal's steerable version \cite{WeilerHS2017} of the FusionNet \cite{QuanHJ16}. It is a UNet \cite{RonnebergerFB15} with added skip connections within the encoder and decoder paths to encourage better gradient flow. We place Gaussian multiplicative dropout with standard deviation 0.1 before and batch normalization after every convolution, and use ReLU nonlinearities directly before each convolution, except on the input. 

For the training set we extract random $100\times100\times5$ voxel patches from the training volume. We reflection pad 10 voxels in the $x$-$y$ plane, and constant pad up to 5 voxels in the the $z$-direction if we sample at the upper or lower image boundaries. We then apply a random elastic distortion in the $x$-$y$-plane, and pass the patches through our group-equivariant FusionNet. We keep our implementation close to the design of Weiler \etal to maintain a close comparison, and do not perform extensive model search. The results are shown in Table \ref{tab:ISBI2012}.
\begin{table}
	\caption{Results for the ISBI 2012 challenge. We have tried to keep our implementation as close as possible to Weiler \etal. Unlike other methods, we perform no post-processing at all unlike Weiler \etal who use a lifting multi-cut \cite{BeierAKH16} post-process, or UNet and Quan \etal who use rotation averaging. Quan also adds an optional median filtering to boost scores. This shows that we can adapt state-of-the-art models to process 3D volumetric data with little change in the competitiveness of the results.}
    \label{tab:ISBI2012}
	\center
	\begin{tabular}{l c c}
    Method								& $V_\text{rand}$	& $V_\text{info}$ \\
    \hline
    UNet \cite{RonnebergerFB15} 		& 0.97276 			& 0.98662	\\
    Quan \etal \cite{QuanHJ16}			& 0.97804			& 0.98995	\\
    Ours								& 0.98018			& 0.98202	\\
    Weiler \etal \cite{WeilerHS2017} 	& 0.98680 			& 0.99144	\\
    IAL MC/LMC							& 0.98792 			& 0.99183
	\end{tabular}
\end{table}

Our results are comparable with other leading methods. Our $V_\text{rand}$ metric is slightly improved over UNet and Quan \etal, but not as good as Weiler \etal, who use a 2D group convolutional neural network approach, with 17 rotations about the $z$-axis and lifting multicut post-process. The leading method uses the lifting multicut method too. Our $V_\text{info}$ metric is not as good as the other methods, but we believe with sufficient model search, and extensive post-processing we could increase this number further. The main point of this experiment, as with the ModelNet10 experiment, was to demonstrate that we could get relatively good performance, without the need for extensive test-time rotation averaging.

\section{Conclusion}
We have presented a 3D convolutional neural network architecture, which is equivariant to right-angle rotations in three dimensions. This relies on an extension of the standard convolution to 3D rotations. On the ModelNet10 classification challenge, we have achieved state-of-the-art for a single model, beating some much larger models, which rely on heavy data augmentation. Since our models are rotation in/equivariant by design, our CNNs need not learn to \emph{overcome} rotations, the way a standard CNN does. In 3D, this is an especially important gain. As a result, our model is positioned to get better generalization with less data, while avoiding the need to perform time-costly rotation averaging at test-time.

Another perspective on our approach is to think of it as global average pooling over rotations, where we expose a new `rotation-dimension.' Without adhering to a defined group, it would be challenging to disentangle or orient a feature space (at any one layer, or across multiple layers) with respect to such a rotation dimension. The trade-off is that we commit to a group and its corresponding CubeNet architecture, to avoid the considerable effort of learning to disentangle pose. 


We leave it to future work to examine whether these models can be generalized to continuous rotations and other challenging transformations, such as scale. There is also the untouched challenge of finding 3D rotation groups, which are not aligned to the Cartesian voxel-grid.

\clearpage

\bibliographystyle{splncs}
\bibliography{egbib}

\begin{thebibliography}{10}

\bibitem{BarnardC91}
Barnard, E., Casasent, D.:
\newblock Invariance and neural nets.
\newblock {IEEE} Trans. Neural Networks \textbf{2}(5) (1991)  498--508

\bibitem{CohenW16}
Cohen, T., Welling, M.:
\newblock Group equivariant convolutional networks.
\newblock In: Proceedings of the 33nd International Conference on Machine
  Learning, {ICML} 2016, New York City, NY, USA, June 19-24, 2016. (2016)
  2990--2999

\bibitem{WorrallGTB17}
Worrall, D.E., Garbin, S.J., Turmukhambetov, D., Brostow, G.J.:
\newblock Harmonic networks: Deep translation and rotation equivariance.
\newblock In: 2017 {IEEE} Conference on Computer Vision and Pattern
  Recognition, {CVPR} 2017, Honolulu, HI, USA, July 21-26, 2017. (2017)
  7168--7177

\bibitem{Chirikjian00}
Chirikjian, G.S.:
\newblock {Engineering Applications of Noncommutative Harmonic Analysis: With
  Emphasis on Rotation and Motion Groups}.
\newblock CRC Press, Abingdon (2000)

\bibitem{CohenW16a}
Cohen, T.S., Welling, M.:
\newblock Steerable cnns.
\newblock CoRR \textbf{abs/1612.08498} (2016)

\bibitem{KondorT18}
Kondor, R., Trivedi, S.:
\newblock On the generalization of equivariance and convolution in neural
  networks to the action of compact groups (2018)

\bibitem{FaselG06}
Fasel, B., Gatica{-}Perez, D.:
\newblock Rotation-invariant neoperceptron.
\newblock In: 18th International Conference on Pattern Recognition {(ICPR}
  2006), 20-24 August 2006, Hong Kong, China. (2006)  336--339

\bibitem{SifreM13}
Sifre, L., Mallat, S.:
\newblock Rotation, scaling and deformation invariant scattering for texture
  discrimination.
\newblock In: 2013 {IEEE} Conference on Computer Vision and Pattern
  Recognition, Portland, OR, USA, June 23-28, 2013. (2013)  1233--1240

\bibitem{OyallonM15}
Oyallon, E., Mallat, S.:
\newblock Deep roto-translation scattering for object classification.
\newblock In: {IEEE} Conference on Computer Vision and Pattern Recognition,
  {CVPR} 2015, Boston, MA, USA, June 7-12, 2015. (2015)  2865--2873

\bibitem{DielemanFK16}
Dieleman, S., Fauw, J.D., Kavukcuoglu, K.:
\newblock Exploiting cyclic symmetry in convolutional neural networks.
\newblock In: Proceedings of the 33nd International Conference on Machine
  Learning, {ICML} 2016, New York City, NY, USA, June 19-24, 2016. (2016)
  1889--1898

\bibitem{GonzalezVT16}
Gonzalez, D.M., Volpi, M., Tuia, D.:
\newblock Learning rotation invariant convolutional filters for texture
  classification.
\newblock In: 23rd International Conference on Pattern Recognition, {ICPR}
  2016, Canc{\'{u}}n, Mexico, December 4-8, 2016. (2016)  2012--2017

\bibitem{LaptevSBP16}
Laptev, D., Savinov, N., Buhmann, J.M., Pollefeys, M.:
\newblock {TI-POOLING:} transformation-invariant pooling for feature learning
  in convolutional neural networks.
\newblock In: 2016 {IEEE} Conference on Computer Vision and Pattern
  Recognition, {CVPR} 2016, Las Vegas, NV, USA, June 27-30, 2016. (2016)
  289--297

\bibitem{GonzalezVKT17}
Gonzalez, D.M., Volpi, M., Komodakis, N., Tuia, D.:
\newblock Rotation equivariant vector field networks.
\newblock In: {IEEE} International Conference on Computer Vision, {ICCV} 2017,
  Venice, Italy, October 22-29, 2017. (2017)  5058--5067

\bibitem{HenriquesV17}
Henriques, J.F., Vedaldi, A.:
\newblock Warped convolutions: Efficient invariance to spatial transformations.
\newblock In: Proceedings of the 34th International Conference on Machine
  Learning, {ICML} 2017, Sydney, NSW, Australia, 6-11 August 2017. (2017)
  1461--1469

\bibitem{Esteves17}
Esteves, C., Allen{-}Blanchette, C., Zhou, X., Daniilidis, K.:
\newblock Polar transformer networks.
\newblock CoRR \textbf{abs/1709.01889} (2017)

\bibitem{ZhouYQJ17}
Zhou, Y., Ye, Q., Qiu, Q., Jiao, J.:
\newblock Oriented response networks.
\newblock In: 2017 {IEEE} Conference on Computer Vision and Pattern
  Recognition, {CVPR} 2017, Honolulu, HI, USA, July 21-26, 2017. (2017)
  4961--4970

\bibitem{WeilerHS2017}
Weiler, M., Hamprecht, F.A., Storath, M.:
\newblock Learning steerable filters for rotation equivariant cnns.
\newblock CoRR \textbf{abs/1711.07289} (2017)

\bibitem{JacobsenBS17}
Jacobsen, J., Brabandere, B.D., Smeulders, A.W.M.:
\newblock Dynamic steerable blocks in deep residual networks.
\newblock CoRR \textbf{abs/1706.00598} (2017)

\bibitem{JacobsenOMS17}
Jacobsen, J.H., Oyallon, E., Mallat, S., Smeulders, A.W.M.:
\newblock Hierarchical attribute cnns.
\newblock In: ICML Workshop on Principled Approaches to Deep Learning. (2017)

\bibitem{ThomasSKYLKR18}
Thomas, N., Smidt, T., Kearnes, S., Yang, L., Li, L., Kohlhoff, K., Riley, P.:
\newblock Tensor field networks: Rotation- and translation-equivariant neural
  networks for 3d point clouds (2018)

\bibitem{LiYLC17}
Li, J., Yang, Z., Liu, H., Cai, D.:
\newblock Deep rotation equivariant network (2017)

\bibitem{Kondor18}
Kondor, R.:
\newblock N-body networks: a covariant hierarchical neural network architecture
  for learning atomic potentials (2018)

\bibitem{CohenGKW18}
Cohen, T.S., Geiger, M., Koehler, J., Welling, M.:
\newblock Spherical cnns (2018)

\bibitem{SimardVLD91}
Simard, P.Y., Victorri, B., LeCun, Y., Denker, J.S.:
\newblock Tangent prop - {A} formalism for specifying selected invariances in
  an adaptive network.
\newblock In: Advances in Neural Information Processing Systems 4, {[NIPS}
  Conference, Denver, Colorado, USA, December 2-5, 1991]. (1991)  895--903

\bibitem{KulkarniWKT15}
Kulkarni, T.D., Whitney, W.F., Kohli, P., Tenenbaum, J.B.:
\newblock Deep convolutional inverse graphics network.
\newblock In: Advances in Neural Information Processing Systems 28: Annual
  Conference on Neural Information Processing Systems 2015, December 7-12,
  2015, Montreal, Quebec, Canada. (2015)  2539--2547

\bibitem{WorrallGTB17b}
Worrall, D.E., Garbin, S.J., Turmukhambetov, D., Brostow, G.J.:
\newblock Interpretable transformations with encoder-decoder networks.
\newblock In: {IEEE} International Conference on Computer Vision, {ICCV} 2017,
  Venice, Italy, October 22-29, 2017. (2017)  5737--5746

\bibitem{SabourFH17}
Sabour, S., Frosst, N., Hinton, G.E.:
\newblock Dynamic routing between capsules.
\newblock In: Advances in Neural Information Processing Systems 30: Annual
  Conference on Neural Information Processing Systems 2017, 4-9 December 2017,
  Long Beach, CA, {USA}. (2017)  3859--3869

\bibitem{CrowleyP84}
Crowley, J.L., Parker, A.C.:
\newblock A representation for shape based on peaks and ridges in the
  difference of low-pass transform.
\newblock {IEEE} Trans. Pattern Anal. Mach. Intell. \textbf{6}(2) (1984)
  156--170

\bibitem{Lindeberg11}
Lindeberg, T.:
\newblock Generalized gaussian scale-space axiomatics comprising linear
  scale-space, affine scale-space and spatio-temporal scale-space.
\newblock Journal of Mathematical Imaging and Vision \textbf{40}(1) (2011)
  36--81

\bibitem{FreemanA91}
Freeman, W.T., Adelson, E.H.:
\newblock The design and use of steerable filters.
\newblock {IEEE} Trans. Pattern Anal. Mach. Intell. \textbf{13}(9) (1991)
  891--906

\bibitem{Lenz90}
Lenz, R.:
\newblock Group Theoretical Methods in Image Processing. Volume 413 of Lecture
  Notes in Computer Science.
\newblock Springer (1990)

\bibitem{SimoncelliFAH92}
Simoncelli, E.P., Freeman, W.T., Adelson, E.H., Heeger, D.J.:
\newblock Shiftable multiscale transforms.
\newblock {IEEE} Trans. Information Theory \textbf{38}(2) (1992)  587--607

\bibitem{Teo98}
Teo, P.C.:
\newblock Theory and Applications of Steerable Functions.
\newblock PhD thesis, Dept. Computer Science, Stanford University (3 1998)

\bibitem{Perona91}
Perona, P.:
\newblock Deformable kernels for early vision.
\newblock In: {IEEE} Computer Society Conference on Computer Vision and Pattern
  Recognition, {CVPR} 1991, 3-6 June, 1991, Lahaina, Maui, Hawaii, {USA}.
  (1991)  222--227

\bibitem{BrunaM13}
Bruna, J., Mallat, S.:
\newblock Invariant scattering convolution networks.
\newblock {IEEE} Trans. Pattern Anal. Mach. Intell. \textbf{35}(8) (2013)
  1872--1886

\bibitem{CotterK17}
Cotter, F., Kingsbury, N.G.:
\newblock Visualizing and improving scattering networks.
\newblock In: 27th {IEEE} International Workshop on Machine Learning for Signal
  Processing, {MLSP} 2017, Tokyo, Japan, September 25-28, 2017. (2017)  1--6

\bibitem{HintonKW11}
Hinton, G.E., Krizhevsky, A., Wang, S.D.:
\newblock Transforming auto-encoders.
\newblock In: Artificial Neural Networks and Machine Learning - {ICANN} 2011 -
  21st International Conference on Artificial Neural Networks, Espoo, Finland,
  June 14-17, 2011, Proceedings, Part {I}. (2011)  44--51

\bibitem{ChenCDHSSA16}
Chen, X., Chen, X., Duan, Y., Houthooft, R., Schulman, J., Sutskever, I.,
  Abbeel, P.:
\newblock Infogan: Interpretable representation learning by information
  maximizing generative adversarial nets.
\newblock In: Advances in Neural Information Processing Systems 29: Annual
  Conference on Neural Information Processing Systems 2016, December 5-10,
  2016, Barcelona, Spain. (2016)  2172--2180

\bibitem{HintonSF18}
Hinton, G.E., Sabour, S., Frosst, N.:
\newblock Matrix capsules with {EM} routing.
\newblock In: International Conference on Learning Representations. (2018)

\bibitem{MaturanaS15}
Maturana, D., Scherer, S.:
\newblock Voxnet: {A} 3d convolutional neural network for real-time object
  recognition.
\newblock In: 2015 {IEEE/RSJ} International Conference on Intelligent Robots
  and Systems, {IROS} 2015, Hamburg, Germany, September 28 - October 2, 2015.
  (2015)  922--928

\bibitem{WuSKYZTX15}
Wu, Z., Song, S., Khosla, A., Yu, F., Zhang, L., Tang, X., Xiao, J.:
\newblock 3d shapenets: {A} deep representation for volumetric shapes.
\newblock In: {IEEE} Conference on Computer Vision and Pattern Recognition,
  {CVPR} 2015, Boston, MA, USA, June 7-12, 2015. (2015)  1912--1920

\bibitem{BrockLRW16}
Brock, A., Lim, T., Ritchie, J.M., Weston, N.:
\newblock Generative and discriminative voxel modeling with convolutional
  neural networks (2016)

\bibitem{SedaghatZB16}
Sedaghat, N., Zolfaghari, M., Brox, T.:
\newblock Orientation-boosted voxel nets for 3d object recognition.
\newblock CoRR \textbf{abs/1604.03351} (2016)

\bibitem{CohenW18}
Cohen, T.S., Geiger, M., K{\"{o}}hler, J., Welling, M.:
\newblock Spherical cnns.
\newblock CoRR \textbf{abs/1801.10130} (2018)

\bibitem{SimonyanZ14a}
Simonyan, K., Zisserman, A.:
\newblock Very deep convolutional networks for large-scale image recognition.
\newblock CoRR \textbf{abs/1409.1556} (2014)

\bibitem{HeZRS15}
He, K., Zhang, X., Ren, S., Sun, J.:
\newblock Delving deep into rectifiers: Surpassing human-level performance on
  imagenet classification.
\newblock In: 2015 {IEEE} International Conference on Computer Vision, {ICCV}
  2015, Santiago, Chile, December 7-13, 2015. (2015)  1026--1034

\bibitem{KingmaB14}
Kingma, D.P., Ba, J.:
\newblock Adam: {A} method for stochastic optimization.
\newblock CoRR \textbf{abs/1412.6980} (2014)

\bibitem{XuT16}
Xu, X., Todorovic, S.:
\newblock Beam search for learning a deep convolutional neural network of 3d
  shapes.
\newblock In: 23rd International Conference on Pattern Recognition, {ICPR}
  2016, Canc{\'{u}}n, Mexico, December 4-8, 2016. (2016)  3506--3511

\bibitem{WuZXFT16}
Wu, J., Zhang, C., Xue, T., Freeman, B., Tenenbaum, J.:
\newblock Learning a probabilistic latent space of object shapes via 3d
  generative-adversarial modeling.
\newblock In: Advances in Neural Information Processing Systems 29: Annual
  Conference on Neural Information Processing Systems 2016, December 5-10,
  2016, Barcelona, Spain. (2016)  82--90

\bibitem{HegdeZ16}
Hegde, V., Zadeh, R.:
\newblock Fusionnet: 3d object classification using multiple data
  representations.
\newblock CoRR \textbf{abs/1607.05695} (2016)

\bibitem{Arganda-Carreras15}
Arganda-Carreras, I., Turaga, S.C., Berger, D.R., CireÅan, D., Giusti, A.,
  Gambardella, L.M., Schmidhuber, J., Laptev, D., Dwivedi, S., Buhmann, J.M.,
  Liu, T., Seyedhosseini, M., Tasdizen, T., Kamentsky, L., Burget, R., Uher,
  V., Tan, X., Sun, C., Pham, T.D., Bas, E., Uzunbas, M.G., Cardona, A.,
  Schindelin, J., Seung, H.S.:
\newblock Crowdsourcing the creation of image segmentation algorithms for
  connectomics.
\newblock Frontiers in Neuroanatomy \textbf{9} (2015)  142

\bibitem{QuanHJ16}
Quan, T.M., Hildebrand, D.G.C., Jeong, W.:
\newblock Fusionnet: {A} deep fully residual convolutional neural network for
  image segmentation in connectomics.
\newblock CoRR \textbf{abs/1612.05360} (2016)

\bibitem{RonnebergerFB15}
Ronneberger, O., Fischer, P., Brox, T.:
\newblock U-net: Convolutional networks for biomedical image segmentation.
\newblock In: Medical Image Computing and Computer-Assisted Intervention -
  {MICCAI} 2015 - 18th International Conference Munich, Germany, October 5 - 9,
  2015, Proceedings, Part {III}. (2015)  234--241

\bibitem{BeierAKH16}
Beier, T., Andres, B., K{\"{o}}the, U., Hamprecht, F.A.:
\newblock An efficient fusion move algorithm for the minimum cost lifted
  multicut problem.
\newblock In: Computer Vision - {ECCV} 2016 - 14th European Conference,
  Amsterdam, The Netherlands, October 11-14, 2016, Proceedings, Part {II}.
  (2016)  715--730

\end{thebibliography}
\end{document}